\documentclass[11pt,twocolumn,twoside]{article}
\usepackage{iccr2024}
\usepackage{placeins}

\begin{document}

\title{Towards Automatic Identification of Missing Tissues using a Geometric-Learning Correspondence Model} 

\author[1,2]{Eliana {Vasquez Osorio}}
\author[1]{Edward Henderson}
\affil[1]{Faculty of Biology, Medicine and Health, University of Manchester, Manchester, UK }
\affil[2]{The Christie NHS Foundation Trust Hospital, Manchester, UK}

\maketitle
\thispagestyle{fancy}

\begin{customabstract} 
Missing tissue presents a big challenge for dose mapping, e.g., in the reirradiation setting. We propose a pipeline to identify missing tissue on intra-patient structure meshes using a previously trained geometric-learning correspondence model. For our application, we relied on the prediction discrepancies between forward and backward correspondences of the input meshes, quantified using a correspondence-based Inverse Consistency Error ($cICE$). We optimised the threshold applied to $cICE$ to identify missing points in a dataset of 35 simulated mandible resections. Our identified threshold, 5.5 mm, produced a balanced accuracy score of 0.883 in the training data, using an ensemble approach. This pipeline produced plausible results for a real case where ~25\% of the mandible was removed after a surgical intervention. The pipeline, however, failed on a more extreme case where ~50\% of the mandible was removed. This is the first time geometric-learning modelling is proposed to identify missing points in corresponding anatomy.
\end{customabstract}

\section{Introduction}

Anatomy of patients undergoing medical treatment often has large `atypical' changes.  In particular, for reirradiation cases in the head and neck region, mandible surgeries due to osteoradionecrosis \cite{Lang2022} are common, resulting in large anatomical mismatches.  The missing tissue between images presents challenges to registration algorithms, with few methods explicitly accounting for them \cite{Nithiananthan2012}. In this study, we propose a pipeline to identify missing tissue on surface meshes of structures using a previously trained geometric-learning correspondence model. This step will enable robust contour-based registrations in particular for estimating cumulative dose metrics in the reirradiation setting. 

\section{Methods}
We used a model previously trained to identify dense point correspondences for head and neck organs at risk \cite{Henderson2023}. 
The correspondence model was trained using an unsupervised approach, where both the geometry of the shapes as well as the imaging information of the planning CT scans were used during training to improve model performance. This model has been shown to be effective for identifying inter-patient correspondences for the brainstem, mandible, parotid glands, spinal cord and submandibular glands. 

At inference time, the model receives two meshes representing the structures, i.e., the source and target meshes, and assigns a vertex on the target mesh to every vertex from the source mesh (defined by $n$ vertices).   More formally, the correspondence between the source and target vertices is represented as the vector $M^{s\rightarrow t}$, of $n$ positions, where the correspondence for a given $i-th$ vertex is $M^{s\rightarrow t}(i) = j$, with $j$ being the index of the corresponding vertex on the target mesh. Note that the model's predictions are not symmetric, as some target vertices may not be assigned to a source vertex.

\subsection{Identification of missing tissue} 
\begin{figure}[b!]
\centering
  \includegraphics[width=0.8\columnwidth]{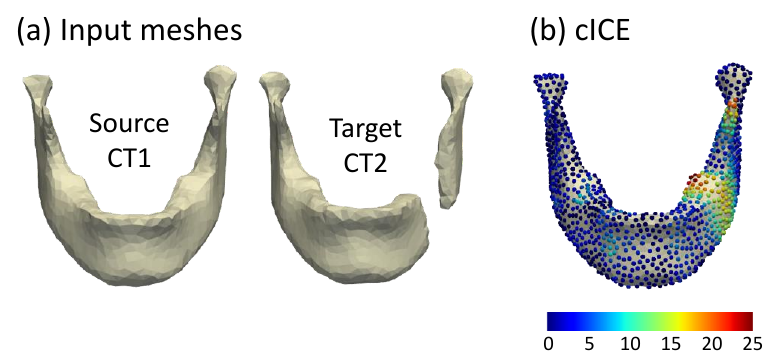}
  \caption{Example of input meshes and the derived $cICE$. Notice that large $cICE$ values are located in regions of point mismatch between the source and target meshes.}
    \label{fig:cICE}
\end{figure}

We propose to identify missing tissue by levering the discrepancy between correspondences predicted by the model when the role of the input meshes are reversed. The first step is to quantify the discrepancies between the correspondence predictions. For this, we estimated the inverse consistency error based on the correspondences, $cICE$ figure \ref{fig:cICE}, using the following equation:
\begin{equation*}
    \label{eq:cICE}
    cICE_i = d(x(i), x(M^{t\rightarrow s}(M^{s\rightarrow t}(i)) )
\end{equation*}

where $d()$ is the Euclidean distance between two points, $x(i)$ is the 3D coordinates of the $i-th$ source vertex, $M^{s\rightarrow t}$ and $M^{t\rightarrow s}$ are the correspondence vectors predicted by the model for the two directions, i.e., the \emph{forward} (source to target) and \emph{backward} (target to source) correspondences. In the ideal case, $M^{t\rightarrow s}(M^{s\rightarrow t}(i))$ is equal to $i$, rendering $cICE(i)$ to zero. As the correspondence model generates discrete predictions, noise is expected on the vertex assignment. Therefore, we applied a median filter to the $cICE$ on the source mesh to mitigate for this effect.

\begin{figure*}[tb]
\centering
  \includegraphics[width=1.0\textwidth]{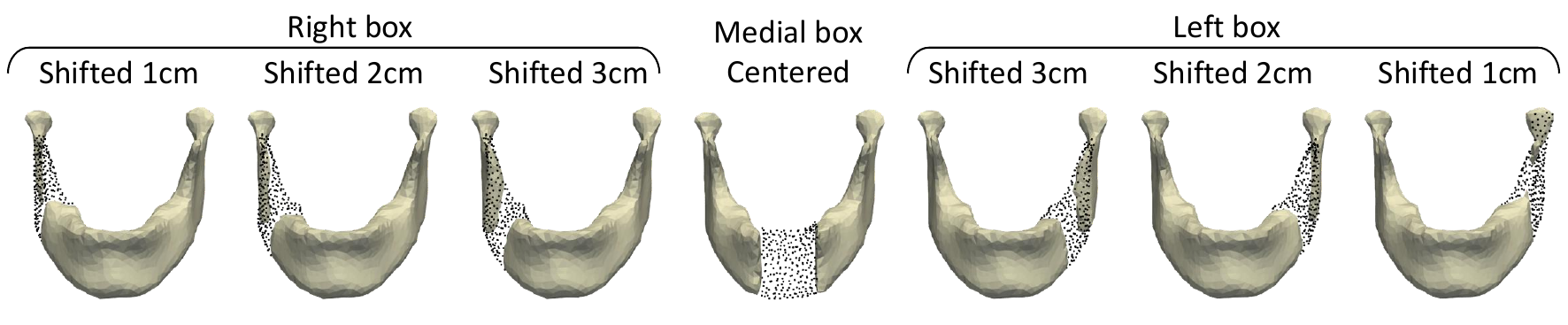}
  \caption{Simulated cuts for the CT2 mandible of a patient. }
    \label{fig:cuts}
\end{figure*}

Our assumption is that points with the largest discrepancies are missing on the target mesh. Therefore, we classify all vertices in the source mesh with a $cICE$ value above a given threshold $t$ as missing in the target mesh. As we applied a median filter in the previous step, we again apply a median filter to `dilate' the selection, in an operation similar to morphological opening filter of binary masks.

\subsection{Pipeline optimisation} 
Mandible contours from 5 patients who underwent reirradiation in the head and neck region were used for this optimisation. Each patient had their mandible delineated on two CT scans: CT1 and CT2, where CT2 was the planning CT scan of the the most recent radiotherapy course. We framed our implementation to predict points present in the CT1 mandible and missing in the CT2 mandible meshes. We optimised two aspects: 1) we compared the performance of an ensemble of multiple correspondence models versus a single model, and 2) we optimised the value of the threshold $t$.

\subsubsection{Ensemble vs single model} As using ensemble approaches have been shown to improve deep-learning performance \cite{Henderson2023ISBI}, we evaluated whether such an approach improved our predictions. For these, we used the predictions of five corresponding models previously generated as part of a 5-fold cross validation\cite{Henderson2023}. We used a simple ensemble approach: majority vote where only points identified as missing by at least three of these single models were kept.

\subsubsection{Threshold optimisation}
We conducted a simulation study to find the optimal value of the threshold $t$ for which we generated simulated mandible resections as training data. We evaluated a range of 19 thresholds, ranging from 1 to 10 mm. We selected the best threshold as the one that maximised the prediction accuracy over all training data. 

\paragraph{Training data generation}   \label{sec:datagen}
We generated training data by artificially cutting the CT2 mandibles using 3cm-wide boxes positioned in different locations, see figure \ref{fig:cuts}. We consistently applied this cutting procedure by using ROI algebra in the scripting interface in RayStation v12.0.100.0 (research) and exporting the generated cut mandibles as masks in MHD format.  These masks were then processed to generate triangular meshes following the same pre-processing steps as presented previously \cite{Henderson2023}.  In total, we generated 35 mesh pairs where between 11\% and 28\% of the mandible volume was removed.

\begin{figure}[h!]
\centering
  \includegraphics[width=0.8\columnwidth]{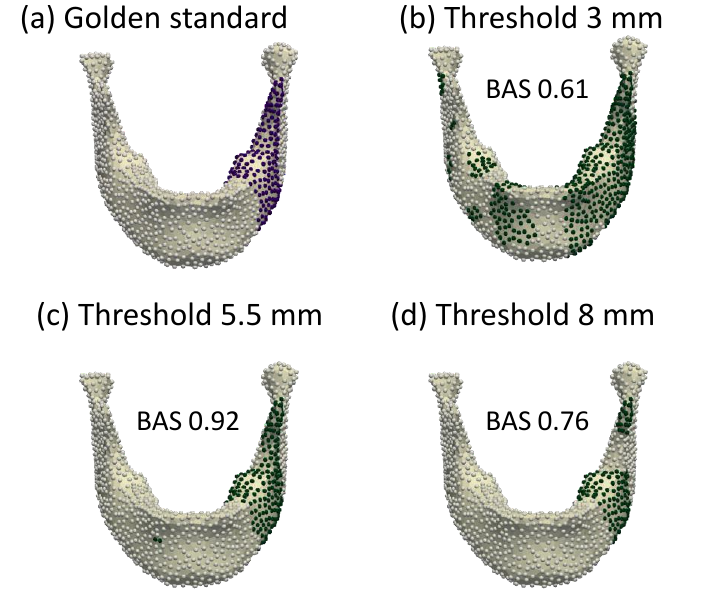}
  \caption{Golden standard for the mesh pair shown in figure \ref{fig:cICE}(a), where the dark points have been identified as missing. (b-d) points identified as missing after applying the proposed pipeline for different thresholds and their balanced accuracy score (BAS). These thresholds are applied to the $cICE$ presented in figure \ref{fig:cICE}(b). }
    \label{fig:BAS}
\end{figure}

\paragraph{Golden standard}  Notice that the ground truth is directly available for the CT2 mandibles via the cutting boxes. However, to evaluate the predictions made by our pipeline, these ground truth needed to be `projected' to the CT1 mandible.  Therefore, we used the forward correspondences for the complete CT2 mandibles to generate the golden standard for missing points on the CT1 mandible mesh: for every mesh pair, a given point was classified as missing if it corresponded to a point within the cutting box. We harmonised the golden standard across the five folds using a majority voting filter, i.e., a given point was classified as missing if it was identified as missing by at least 3 single models. See figure \ref{fig:BAS}(a) for an example.

\begin{figure*}[t]
\centering
  \includegraphics[width=0.9\textwidth, trim={0 1cm 0 0},clip]{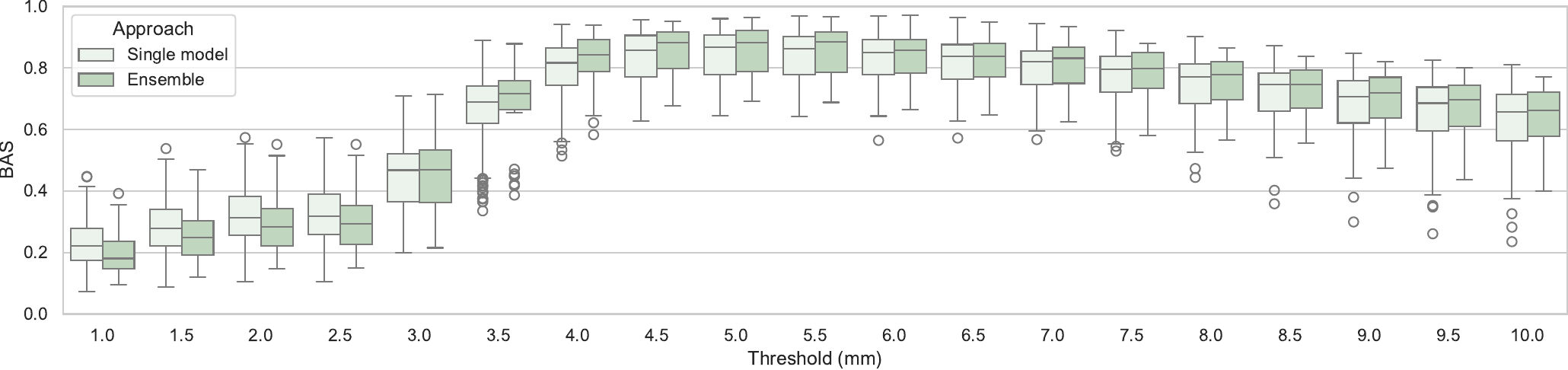}
  
  \includegraphics[width=0.9\textwidth]{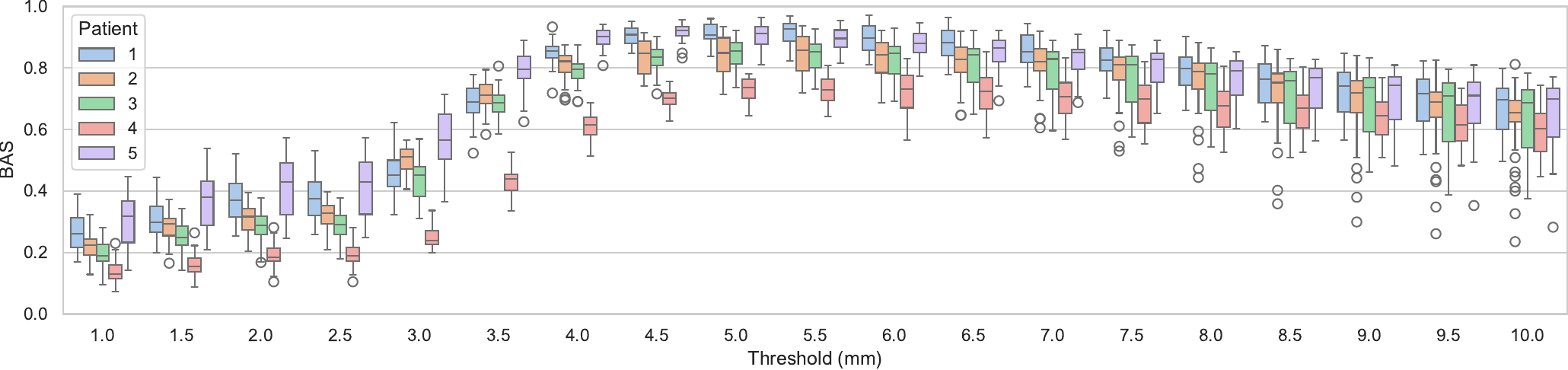}
  \caption{Balanced accuracy score (BAS) for all simulations and tested thresholds, discriminating by single model vs ensemble approach (top) and by patient (bottom).}
    \label{fig:boxplots}
\end{figure*}

\paragraph{Prediction evaluation} 
We evaluated the prediction of missing points for different thresholds against the golden standard using an accuracy score.  As we expected to have classifications with imbalanced class distributions, e.g., most points are not missing in CT2, we used the balanced accuracy score (BAS). BAS has been demonstrated to be robust to imbalanced  class distributions \cite{Brodersen2010} and it is defined as the average of recall obtained on each class, figure \ref{fig:BAS}.  We used the implementation provided in scikit-learn v1.3.2.

\subsubsection{Qualitative evaluation of generalisation}
We applied the most optimal pipeline and threshold to two extra cases where the patients underwent mandible surgery in between the two radiotherapy courses due to osteoradionecrosis. These patients lost ~25\% and ~50\% of the mandible during the surgical intervention, respectively.

\section{Results}

Figure \ref{fig:boxplots} shows the performance of the pipeline. BAS above 0.8 are observed for both approaches at thresholds between 4.0 and 7.0 mm, with the largest median being 0.867 for the single model approach at 5.0 mm and 0.883 for the ensemble approach at 5.5 mm. The performance of the classification pipeline was different for some patients, with results for patient 4 systematically lower for most thresholds. The pipeline performed best for patients 5 and 1, followed by patients 2 and 3. 

\begin{figure}[h]
\centering
  \includegraphics[width=0.8\columnwidth]{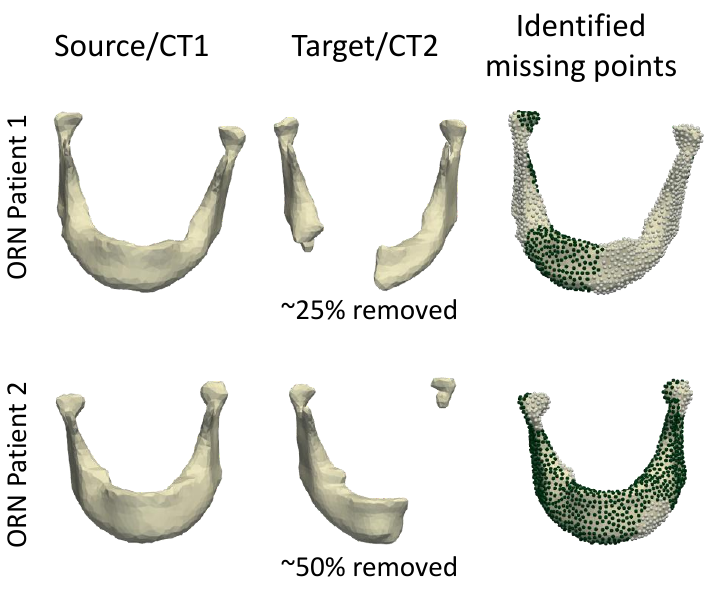}
  \caption{Performance of the ensemble pipeline using the found optimal threshold of 5.5 mm to data of two patients who underwent mandible surgery due to osteoradionecrosis after their first radiotherapy course.}
    \label{fig:orn}
\end{figure}

\begin{figure}[h]
\centering
  \includegraphics[width=1.0\columnwidth]{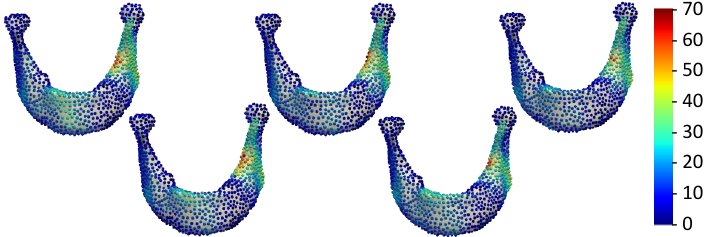}
  \caption{$cICE$ for all folds for ORN patient 2. Note the pattern similarity with $cICE$ presented in figure \ref{fig:cICE}}
    \label{fig:orn_cice}
\end{figure}

Figure \ref{fig:orn} shows the results of applying the ensemble pipeline with threshold 5.5 mm on two extra patients who underwent surgical intervention due to mandible osteoradionecrosis after their first radiotherapy course. The results on ORN patient 1 are plausible with the largest region identifying the most anterior part of the mandible.  Two small regions, up in the mandible head and the coronoid process were also identified.  The prediction for ORN patient 2 was not successful, as most of the points were identified as missing. However, the calculated $cICE$ followed a similar pattern as other mesh pairs but with a much higher magnitude, compare figure \ref{fig:orn_cice} and \ref{fig:cICE}. 

\section{Discussion}
In this study we proposed, implemented and optimised a pipeline to identify missing tissue on intra-patient organ meshes based on a previously trained geometric-learning correspondence model, previously shown to be effective at identifying inter-patient correspondences for head and neck organs. For our application, we relied on the prediction discrepancies between the forward and backward correspondences quantified via correspondence-based inverse consistency error or $cICE$, where points with $cICE$ above a given threshold were identified as missing.  We  optimised this threshold in a dataset of 35 simulated mandible resections by evaluating the balanced accuracy score (BAS) between the golden standard and the predictions. Our identified threshold, 5.5 mm, produced a BAS of 0.883 in the training data using an ensemble approach.  This pipeline was then shown to produce plausible results for a real case of a patient where ~25\% of their mandible was missing after a surgical intervention.  The pipeline, however, failed to identify missing points on a more extreme case, where ~50\% of the mandible was removed.

This is the first time that geometric-learning correspondence models are used to identify missing vertices on organ meshes. A recent related work using a deep-learning approach is the method proposed by Andresen and colleagues \cite{Andresen2022} where they use a convolutional neural network to jointly perform non-rigid registration and segmentation of non-corresponding anatomy. They applied their proposed network to inter-patient MR scans with artificially added stroke lesions. The reported Jaccard indices\footnote{Overlap-based contour similarity measure similar to the Dice Similarity Coefficient} ranging between 0.59 and 0.62 highlights the need to find better solutions for identifying missing tissues. 

Comparing our method to more traditional approaches within radiotherapy, the work published by Vasquez Osorio and colleagues \cite{VasquezOsorio2012} is relevant. The authors used an adapted version of the thin-plate spline robust-point matching algorithm to register automatically segmented features between livers contoured on CT and MR. They demonstrated that outliers present in both datasets could be identified by using fuzzy correspondences, even for large numbers (such as 55\% of input points). Similarly, the work by Myronenko and Song \cite{Myronenko2010} has been shown to generate robust registration results in the presence of noise and missing points, which could then be indirectly used to identify outliers/missing points. However, a common disadvantage is the long computational time. 

This pipeline was shown to produce plausible results for a case where ~25\% of their mandible was missing. However, the pipeline failed in the second, more extreme case, where ~50\% of the mandible was removed. Several factors can be related to this failure. First, the extreme case was out of distribution for the training dataset (11\%-28\% of the mandibles removed). Therefore, the 5.5 mm could be a very conservative value for this case, especially as the $cICE$ reached 70 mm, see Figure \ref{fig:orn_cice}. Defining an adaptive thresholds, informed by the difference in areas or volumes between the input meshes, and evaluating the pipeline in other `softer' organs are areas of future development. 

Another limitation was the use of an  model originally trained for a different purpose. The correspondence model was trained with complete organs which always consisted of a single connected mesh. Designing a correspondence model that explicitly includes classification of missing points is also a promising avenue of research. In this case, the procedure described in section \ref{sec:datagen} could be used as data augmentation strategy. Additionally, using volumetric meshes (composed of tetrahedra) rather than surface meshes (composed of triangles) and incorporating imaging at inference time may also provide extra robustness.

\paragraph{Conclusion}
We have demonstrated how a correspondence model and a post-processing pipeline can be used to identify missing tissue on surface meshes of structures of the same patient using a previously trained geometric-learning correspondence model. Further work to enable its use on more extreme cases is required. We are paving the way to robustly utilise contour-based registrations for radiotherapy applications, such as cumulative dose estimations in the reirradiation setting.

\printbibliography

@InProceedings{Henderson2023,
author="Henderson, Edward G. A.
and van Herk, Marcel
and Green, Andrew F.
and Vasquez Osorio, Eliana M.",
title="Unsupervised Correspondence with Combined Geometric Learning and Imaging for Radiotherapy Applications",
booktitle="Shape in Medical Imaging",
year="2023",
publisher="Springer Nature Switzerland",
address="Cham",
pages="75--89",
isbn="978-3-031-46914-5",
doi="10.1007/978-3-031-46914-5_7"
}

@INPROCEEDINGS{Brodersen2010,
  author={Brodersen, Kay Henning and Ong, Cheng Soon and Stephan, Klaas Enno and Buhmann, Joachim M.},
  booktitle={2010 20th International Conference on Pattern Recognition}, 
  title={The Balanced Accuracy and Its Posterior Distribution}, 
  year={2010},
  volume={},
  number={},
  pages={3121-3124},
  doi={10.1109/ICPR.2010.764}}

@article{Andresen2022,
author = {Andresen, Julia and Kepp, Timo and Ehrhardt, Jan and Burchard, Claus and Roider, Johann and Handels, Heinz},
year = {2022},
month = {03},
pages = {},
title = {Deep learning-based simultaneous registration and unsupervised non-correspondence segmentation of medical images with pathologies},
volume = {17},
journal = {International Journal of Computer Assisted Radiology and Surgery},
doi = {10.1007/s11548-022-02577-4}
}

@article{VasquezOsorio2012,
title = "Accurate CT/MR vessel-guided nonrigid registration of largely deformed livers",
author = "{Vasquez Osorio}, Eliana and Mischa Hoogeman and {Mendez Romero}, Alejandra and Piotr Wielopolski and AG Zolnay and Ben Heijmen",
year = "2012",
doi = "10.1118/1.3701779",
volume = "39",
pages = "2463--2477",
journal = "Medical Physics",
number = "5"
}

@ARTICLE{Myronenko2010,
  author={Myronenko, Andriy and Song, Xubo},
  journal={IEEE Transactions on Pattern Analysis and Machine Intelligence}, 
  title={Point Set Registration: Coherent Point Drift}, 
  year={2010},
  volume={32},
  number={12},
  pages={2262-2275},
  doi={10.1109/TPAMI.2010.46}}

@article{Nithiananthan2012,
author = {Nithiananthan, Sajendra and Schafer, Sebastian and Mirota, Daniel J. and Stayman, J. Webster and Zbijewski, Wojciech and Reh, Douglas D. and Gallia, Gary L. and Siewerdsen, Jeffrey H.},
title = {Extra-dimensional Demons: A method for incorporating missing tissue in deformable image registration},
journal = {Medical Physics},
volume = {39},
number = {9},
pages = {5718-5731},
doi = {https://doi.org/10.1118/1.4747270},
year = {2012}
}

@INPROCEEDINGS{Henderson2023ISBI,
  author={Henderson, Edward G. A. and van Herk, Marcel and Osorio, Eliana M. Vasquez},
  booktitle={2023 IEEE 20th International Symposium on Biomedical Imaging (ISBI)}, 
  title={The Impact of Training Dataset Size and Ensemble Inference Strategies on Head and Neck Auto-Segmentation}, 
  year={2023},
  volume={},
  number={},
  pages={1-4},
  doi={10.1109/ISBI53787.2023.10230826}}

@article {Lang2022,
	Title = {Frequency of osteoradionecrosis of the lower jaw after radiotherapy of oral cancer patients correlated with dosimetric parameters and other risk factors},
	Author = {Lang, Kristin and Held, Thomas and Meixner, Eva and Tonndorf-Martini, Eric and Ristow, Oliver and Moratin, Julius and Bougatf, Nina and Freudlsperger, Christian and Debus, Jürgen and Adeberg, Sebastian},
	DOI = {10.1186/s13005-022-00311-8},
	Number = {1},
	Volume = {18},
	Month = {February},
	Year = {2022},
	Journal = {Head \& face medicine},
	ISSN = {1746-160X},
	Pages = {7}
}

\end{document}